# Sur le statut référentiel des entités nommées

Thierry Poibeau

Laboratoire d'Informatique de Paris Nord – CNRS et Université Paris 13
99, av. J.-B. Clément – F-93430 Villetaneuse
thierry.poibeau@lipn.univ-paris13.fr

**Résumé** Nous montrons dans cet article qu'une même entité peut être désignée de multiples façons et que les noms désignant ces entités sont par nature polysémiques. L'analyse ne peut donc se limiter à une tentative de résolution de la référence mais doit mettre en évidence les possibilités de nommage s'appuyant essentiellement sur deux opérations de nature linguistique : la synecdoque et la métonymie. Nous présentons enfin une modélisation permettant de rendre explicite les différentes désignations en discours, en unifiant le mode de représentation des connaissances linguistiques et des connaissances sur le monde.

**Abstract** We show in this paper that, on the one hand, named entities can be designated using different denominations and that, on the second hand, names denoting named entities are polysemous. The analysis cannot be limited to reference resolution but should take into account naming strategies, which are mainly based on two linguistic operations: synecdoche and metonymy. Lastly, we present a model that explicitly represents the different denominations in discourse, unifying the way to represent linguistic knowledge and world knowledge.

## 1 Introduction

Les entités nommées désignent les noms de personnes, de lieux, d'organisations mais aussi les dates ou les unités monétaires. Alors que ces éléments ont longtemps été délaissés par les systèmes de traitement automatique des langues, le renouveau lié au travail sur corpus a révélé qu'il s'agissait en fait d'éléments majeurs pour l'analyse. Les conférences en extraction d'information (*Message Understanding Conferences*, cf. MUC-7, 1998) ont mis en avant plusieurs tâches génériques, au premier rang desquelles l'analyse des entités nommées. Il s'agit en fait d'une double tâche : d'une part une tâche de reconnaissance des séquences pertinentes, d'autre part une tâche de typage des séquences ainsi reconnues en fonction d'une ontologie pré-établie. Sur des textes de type journalistique, les systèmes obtiennent généralement d'assez bons scores, avec un taux combiné de rappel et de précision supérieur à 0,90.

Les entités nommées sont particulièrement importantes pour l'accès au contenu du document car elles forment les briques élémentaires sur lesquelles repose l'analyse. Les entités sont généralement considérées comme directement référentielles : ce sont les désignateurs rigides de Kripke (1982) qui font référence aux objets du monde, organisés en ontologie. Nous montrons ici les limites de cette approche et nous contestons la vue traditionnelle qui rend compte d'une façon très simplifiée de la complexité de la langue. Nous proposons quelques pistes pour mieux prendre en compte le sémantisme complexe des entités.

Nous présentons dans un premier temps les systèmes classiques de repérage d'entités nommées. Ceux-ci reposent sur un ensemble de règles permettant de typer les séquences pertinentes d'après un ensemble de catégories prédéfinies. Nous montrons ensuite les limites de cette approche : la plupart des entités sont polysémiques et leur analyse dépend étroitement du contexte. Partant de ce constat, l'analyse mise en œuvre ne peut qu'être dynamique et refléter les effets de sens en contexte. Nous proposons enfin, dans une dernière partie, une modélisation à base de structures de traits permettant de mieux rendre compte des phénomènes en jeu.

## 2   Systèmes de repérage et de catégorisation des entités nommées

Les entités nommées sont traditionnellement typées suivant une hiérarchie pré-établie. Nous examinons ce type de hiérarchie et les problèmes posés par les textes.

### 2.1   Hiérarchies de types d'entités

Sous l'influence des conférences américaines d'évaluation MUC, les travaux en extraction d'entités nommées ont traditionnellement été effectués sur des textes journalistiques ou des dépêches d'agence. L'identification des entités nommées inclut trois types d'expressions :

- ENAMEX : les expressions de noms propres incluant les noms de personnes, de lieux et d'organisations.
- TIMEX : les expressions temporelles comme les dates et les heures.
- NUMEX : les expressions numériques telles que les expressions monétaires et les pourcentages.

Voici par exemple la hiérarchie de types définie par le LIMSI pour le système QALC (Ferret *et al.*, 2001) et calqué sur la hiérarchie définie pour MUC (Grishman et Sundheim, 1995).

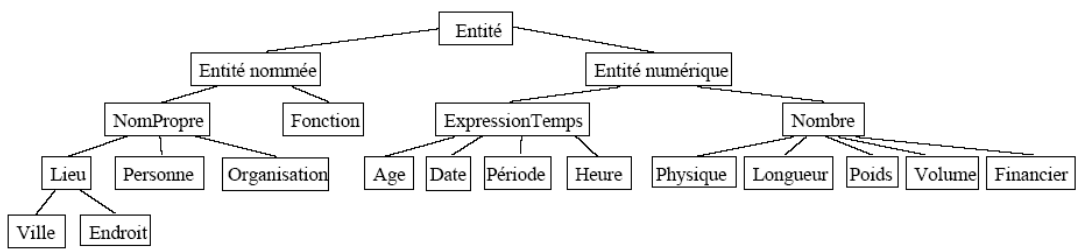

Les hiérarchies comportent ainsi, pour les plus simples, une douzaine de types de base (feuilles de la hiérarchie) mais ont souvent besoin d'être étendues pour couvrir de nouveaux besoins (de nouvelles tâches, de nouveaux domaines). Il n'est ainsi pas rare de faire face à des hiérarchies de plus de 200 éléments (Sekine, 2004).

## 2.2 Repérage et classification des entités nommées

De nombreux travaux ont porté sur l'identification des noms propres dans des textes journalistiques, notamment les *Message Understanding Conferences* (MUC7, 1998). La reconnaissance des entités nommées à partir de textes écrits est actuellement la tâche d'extraction d'information qui obtient les meilleures performances. Les performances sont mesurées en utilisant des mesures classiques comme P&R, fondée sur un taux combiné de précision et de rappel suivant la règle suivante : $P\&R = \frac{2*PRECISION*RAPPEL}{PRECISION+RAPPEL}$. Les taux obtenus sont comparables à ceux des humains, de l'ordre de 0,90 P&R sur des dépêches journalistiques.

Deux grandes approches sont généralement suivies pour leur identification : une approche linguistique « de surface » et une approche probabiliste. L'approche linguistique est fondée sur la description syntaxique et lexicale des syntagmes recherchés. Des règles de grammaire utilisent des marqueurs lexicaux (ex. *Mr* pour *Mister* ou *Inc.* pour *Incorporated*), des dictionnaires de noms propres et des dictionnaires de la langue générale (essentiellement pour repérer les mots inconnus) sont utilisés pour repérer et typer les syntagmes intéressants (Aberdeen *et al.*, 1995 ; Grishman, 1995 ; Appelt et Martin, 1999). De son côté, l'approche probabiliste utilise un modèle de langage entraîné sur de larges corpus de textes pré-étiquetés. Cette approche est particulièrement robuste lorsque les textes sont bruités, c'est pourquoi la grande majorité des systèmes dédiés à l'oral adoptent une telle approche (Kubala, 1999).

# 3 Difficultés et limites de la catégorisation

Nous montrons dans cette partie les limites de l'approche traditionnelle. Nous montrons notamment la polysémie fréquente des entités nommées et les conséquences pour l'analyse automatique.

## 3.1 Polysémie des entités nommées

La catégorisation des entités repose en grande partie sur l'hypothèse que les entités sont référentielles et peuvent donc recevoir un type rendant compte de leur référent. Cette analyse est le plus souvent erronée comme on peut aisément le montrer, les entités nommées référant le plus souvent à plusieurs classes (Leroy, 2004). Les dates se confondent ainsi fréquemment avec des événements :

> *Le 11 septembre 2001 a représenté un tournant dans l'histoire américaine. (Elie Wiesel, site www.france-amerique.com)*

Il est parfois difficile de classer les noms d'organisations, qui peuvent être catégorisés comme institution, communauté d'individus ou encore bâtiment.

> *Le journal télévisé a eu lieu hier en direct de l'ONU.*
> *L'ONU était en grève hier.*
> *L'ONU a fêté ses 50 ans.*
> *L'ONU n'acceptera pas une attaque frontale de l'Irak (forum du Monde)*

On retrouve le même phénomène pour les noms de lieu :

*L'Europe veut garder la tête du FMI. (Libération, 10 mars 2004)*

Les noms de personnes sont encore plus labiles. Nous laisserons de côté les exemples où le nom de personne est en fait devenu la désignation d'une entreprise, d'un stade ou d'un lieu quelconque.

*Une rencontre d'un niveau technique assez médiocre à l'Abbé Deschamps. (stade d'Auxerre, Journal L'équipe)*

Même sans prendre en compte ces phénomènes de transfert de sens, les exemples de catégorisation changeante sont légion. Un nom de personne peut référer à une œuvre, à un objet ou à tout autre élément ayant un lien direct ou non avec la personne.

*J'ai tout Chirac sur l'étagère*
*Pierre est garé en face. (cf. Cadiot et Visetti 2001, p. 167)*

Il existe par ailleurs des phénomènes plus rares mais bien attestés où un nom propre, particulièrement un nom de la personne, ne renvoie pas au référent traditionnel. C'est par exemple le cas de l'antonomase, largement étudiée par S. Leroy (2003) : *mon oncle est un vrai Harpagon* ne réfère pas à un personnage réel s'appelant Harpagon.

On voit à travers ces exemples rapides que les entités nommées ne se comportent pas fondamentalement différemment des autres unités linguistiques. L'exemple le plus connu est certainement celui du *Prix Goncourt* introduit initialement par Kayser (1988). Celui-ci distingue plusieurs sens différents, suivant qu'il s'agisse du prix, de sa valeur monétaire, du livre qui a obtenu le prix, de l'institution ou encore de l'auteur.

## 3.2 Référentialité des entités nommées

Nous retrouvons sous ces thèmes déjà évoqués à de multiples reprises dans la littérature, le questionnement habituel sur la notion de polysémie. Comment traiter ces cas ? Leurs sens sont-ils mutuellement incompatibles ou peut-on définir une théorie unificatrice du sens rendant compte de ces multiples emplois ?

Les entités nommées constituent à cet égard une classe de syntagmes proche des termes. De nombreux travaux étrangers ne font d'ailleurs le plus souvent pas de différences entre ces éléments (Collier *et al.*, 2000), d'autant que certains domaines contribuent à rendre floues les distinctions qui semblent classiques par ailleurs. Dans une phrase issue d'un corpus de biologie comme *Abd-a interagit avec trx*, a-t-on affaire à des entités nommées ou à des termes ? *Abd-a* constitue-t-il un nom de gènes ou une classe ? Les réponses ne sont pas toujours claires. Le propre des entités nommées est malgré tout d'être référentielles. C'est essentiellement ce trait qui définit leur nature et qui les distingue des termes. Leurs propriétés référentielles ont été bien étudiées, depuis Frege (1892) et Kripke (notion de désignateur rigide, cf. Kripke, 1982), jusqu'aux travaux récents dans des cadres linguistiques divers (Charolles, 2002).

On remarque cependant la même variation au niveau des entités nommées qu'au niveau des termes. Il existe de multiples façons de désigner une personne ou un objet, il n'y a pas de nom unique et inévitable (cf. l'exemple de Frege autour de l'*étoile du soir* et *l'étoile du matin* qui désignent toutes deux Vénus …). On peut dès lors s'interroger sur la nature référentielle de l'entité nommée employée en discours. Il ne s'agit bien évidemment pas de nier les possibilités référentielles de l'entité nommée mais il est permis de se demander si l'emploi d'une entité nommée en discours ne laisse pas de côté son statut référentiel : celui-ci serait en quelque sorte dans un état latent. Concrètement, cela signifie qu'en discours il n'y a pas de réelle différence entre une affirmation portant sur une entité, sur un terme ou sur un autre groupe nominal. Comme le rappelle Charolles (2002, p. 7-10), dans la littérature linguistique, les auteurs distinguent la dénotation de la référence. La dénotation « prend acte du fait que le signifié (des signes linguistiques) exprime un concept muni d'une certaine intension (i.e. un ensemble d'attributs caractéristiques en propre) et que cette intension délimite virtuellement son extension (c'est-à-dire une classe d'êtres satisfaisants ces attributs) ». L'acte de référence est quant à lui « un accord non entre deux pensées (…) mais entre deux pensées à propos de quelque chose et cela par le biais de la production, en contexte, d'une expression référentielle ». Nous reprenons à notre compte l'essentiel de ces définitions mais à l'inverse de Charolles, nous soutenons que l'on peut alors parler d'opération linguistique pour la dénotation, contrairement à la référence qui ressortit à une logique visant à appairer niveau linguistique et niveau extra-linguistique.

En discours, l'aspect dénotationnel est primordial, passant bien avant l'aspect référentiel. Cette distinction n'est pas seulement d'ordre philosophique : il s'agit d'une dimension essentielle de l'interprétation du texte dans la mesure où la dénotation, à l'inverse de la référence, peut supporter une sous-spécification par défaut[1]. Ainsi, dans l'exemple de biologie ci-dessus, la résolution du fait que *Abd-a* réfère à un gène particulier ou à une classe n'a pas lieu d'être dans la mesure où cela est inutile pour l'interprétation de la phrase. Le fait que l'*étoile du soir* et l'*étoile du matin* désignent la même planète n'est pas fondamental au niveau linguistique mais est en revanche essentiel pour assigner une valeur de vérité aux assertions émises sur ces objets[2]. On rejoint ici le point de vue de Searle (1983) pour qui « l'acte de référence, le processus de référenciation », inclut une dimension externe à la langue « à savoir que l'entité visé par cet acte, la chose (au sens le plus général du terme) à laquelle il renvoie, existe au-delà de ce que le locuteur en dit, en dehors de son esprit et de celui de celles ou ceux à qui il s'adresse ». (Charolles, 2002, p. 38)

## 3.3 Catégorisation à partir d'une ontologie

La plupart des auteurs qui se sont intéressés à la question du traitement automatique de la métonymie ou de la synecdoque proposent des approches fondées sur des ressources générales comme Wordnet (cf. l'atelier *The lexicon and figurative language* durant ACL 2003 [Wellington, 2003]). Il est en effet difficile d'acquérir directement à partir du corpus des informations comme une décomposition des objets en relation *partie-tout*. L'analogie peut jouer un rôle et permettre le repérage de relations de dépendance spécifiques (voir [Nehaniv, 1999] ou [Lepage, 2003]) mais ceci reste toutefois marginal dans la plupart des cas.

Les exemples et les remarques de la section précédente semblent toutefois aller dans le sens de la théorie défendue par Cadiot et Visetti (2001), qui contestent l'existence préalable d'une ontologie, notamment pour les groupes nominaux déterminés et plus généralement, pour les unités dites référentielles. La notion de « facette » (Cruse, 2004) rend partiellement compte de l'aspect polysémique des entités envisagées mais plusieurs auteurs ont souligné que les facettes ne rendent pas compte des liens entre les différents sens envisagées. Cadiot et Visetti montrent bien qu'il s'agit essentiellement de phénomènes de synecdoque et de métonymie, qui ruinent toute tentative directement référentielle mais qui appellent plutôt une analyse dynamique par « profilage » de sens en fonction du contexte, pour reprendre une partie de la terminologie employée par les deux auteurs (sur ces questions, voir aussi Fass, 1988). Chibout (1999) donne une typologie claire de plusieurs figures de style impliquant des liens sémantiques complexes entre unités linguistiques mises en jeu. Une modélisation par un ensemble de traits activables en contexte semble particulièrement approprié.

---

[1] On retrouve ici la notion de profondeur variable de Kayser et Coulon [1981], que nous interprétons ainsi : l'analyse n'a pas à se préoccuper d'un certain nombre de traits sémantiques latents tant que ceux-ci ne sont pas activés de manière explicite. D'une manière plus globale, il n'est pas toujours nécessaire de chercher à désambiguïser finement le rôle de l'entité suivant la tâche ou l'application visée.

[2] Les cas où l'analyse de la référence est obligatoire sont somme toute relativement rares pour les applications de traitement automatique des langues. Le cadre privilégié est la commande de système, la communication avec des robots ou tout autre cadre où la parole entraîne une action très directe sur le monde.

Plusieurs auteurs ont proposé des modèles informatiques visant à résoudre des problèmes relativement similaires aux nôtres. La plupart se situent dans le cadre du dialogue homme-machine et s'en tiennent à identifier les problèmes de référence qui se posent lors de l'analyse (Salmon-Alt, 2001). Citons toutefois l'étude de Pospescu-Belis et al. (1999) qui présente un modèle de résolution des anaphores fondé sur le modèle des représentations mentales. L'auteur évoque la notion de point de vue et la variabilité de la référence dans un cadre conversationnel. Il faut d'ailleurs souligner le rôle de la tâche pour ce type d'analyse.

Il semble toutefois difficile de se passer de toute catégorisation préalable, de se fonder uniquement sur une approche constructiviste, voire herméneutique, à partir d'unité de sens activables (ou non) en contexte[3]. Les catégories prédéterminées souffrent des mêmes défauts que les facettes de Cruse : l'analyse ne dit rien des rapports entre date et événement, et, plus généralement, n'explique pas le continuum qui existe entre les différents aspects sémantiques des entités en contexte. Mais au moins une telle analyse ne préjuge pas de la sémantique de l'entité en contexte, elle laisse ouverte une liste de possibles. On laissera à la sémantique lexicale, voire infra lexicale, le soin de mettre au jour les liens entre traits activables au sein du mot.

## 4    Proposition de modélisation

Nous gardons de ce qui précède l'idée d'hypothèses construites en parallèle, pouvant être activées ou non en contexte. Nous nous inspirons des structures de traits à la DATR (Evans et Gazdar, 1996) et du lexique génératif (Pustejovsky, 1995) pour coder les différents traits activables pour les entités.

### 4.1    Modélisation sous forme de structures de traits

La hiérarchie de types définie dans le cadre des conférences MUC et présentée dans la première partie de cette étude a prouvé son efficacité. Il nous semble donc pertinent de nous appuyer sur celle-ci pour définir des types de base.

La plupart des propositions qui ont par la suite été faites pour raffiner cette hiérarchie ou d'autres catégorisations ayant une granularité comparable (Sekine, 2004) ont été confrontées

---

[3] Une proposition allant toutefois dans ce sens est celle de Sabah (1996) qui propose un modèle informatisé appelé « carnet d'esquisses ». On retrouve sous ce terme une vue dynamique de la sémantique en train de se construire de manière dynamique, en fonction du co-texte et du contexte.

aux problèmes de polysémie soulevés ci-dessus. Pour reprendre un exemple déjà entrevu : comment faire un choix parmi les différents sens du mot ONU, a-t-on affaire à l'institution, au bâtiment ou à l'ensemble des personnes qui composent l'organisme ? Il s'agit essentiellement d'un mécanisme de focalisation permettant de mettre en avant un aspect de l'entité en contexte. Nous proposons donc d'adjoindre un trait *saillance* aux entités afin de mettre en avant de manière explicite cette focalisation qui peut changer de manière dynamique en fonction du contexte. Reprenons quelques uns des exemples déjà traités :

*L'ONU n'acceptera pas une telle décision.*
```
Entity{
    Lexical_unit=ONU;
    Sem{
        Type=organization;
        Focalisation=diplomatic_org; }
}
```

L'ONU désigne dans ce cadre l'organisation politique. Cette information peut être repérée à partir d'informations sur le monde, sur le rôle de l'ONU sur la scène diplomatique et sur l'analyse du groupe verbal qui suit l'entité. On peut rapprocher cette information du rôle *télique* défini par Pustejovsky (1995) : le propre de l'ONU, c'est de prendre et de faire appliquer des décisions d'ordre diplomatique.

*Le journal télévisé a eu lieu en direct de l'ONU.*
```
Entity{
    Lexical_unit=ONU;
    Sem{
        Type=organization;
        Focalisation=location; }
}
```
Ici le verbe *avoir lieu* fait clairement allusion à un processus en train de se dérouler dans un lieu donné. La focalisation porte donc sur la localisation de l'événement.

*L'ONU était en grève hier.*
```
Entity{
    Lexical_unit=ONU;
    Sem{
        Type=organization;
        Focalisation=human_org; }
}
```
Le même phénomène opère ici. La notion de *grève* active la notion d'organisation composée d'individus (*human_org*). Le trait de focalisation est donc mis à jour. L'exemple qui suit est à cet égard moins clair :

*L'ONU a fêté ses 50 ans.*
```
Entity{
    Lexical_unit=ONU;
    Sem{
        Type=organization;
        Focalisation=none; }
}
```
L'auteur prête ici des traits humains à l'institution. Il ne s'agit plus d'une simple synecdoque mais d'un phénomène métaphorique, en tant que tel plus difficile à analyser automatiquement. D'un côté, le verbe *fêter* invite à voir une communauté d'individus dans le sujet. Mais c'est bien l'institution qui a 50 ans. Une focalisation sur la communauté d'individus serait quelque peu abusive dans ce cas. Peut-être faut-il s'en tenir à une certaine sous-spécification au niveau de l'analyse automatique : l'institution et les individus qui la composent forment ici une unité syncrétique qui échappe partiellement à l'analyse.

## 4.2 Du formulaire d'entités à la résolution des anaphores nominales

L'analyse des entités suivant le modèle que nous proposons demande à avoir accès à des informations sur les entités. Ces informations peuvent être qualifiées de connaissances sur le monde ou encore de connaissances de sens commun. La façon d'encoder ces informations doit être générique mais il semble encore hors de portée de développer des systèmes à large couverture ayant une telle masse de connaissances sur le monde. Le plus grand projet connu allant dans ce sens est le projet CYC de Doug Lenat (Lenat et Guha, 1990) mais les articles de l'auteur ont depuis révélé que la tâche était sans doute hors d'atteinte et les réalisations ont jusqu'à maintenant été peu concluantes (Lenat, 2001).

L'analyse des entités demande une modélisation fine du domaine visé. Une tentative allant dans ce sens est celle des conférences d'extraction d'information à partir de textes MUC (Message Understanding Conferences). Plusieurs tâches génériques ont été définies dans le cadre de ces conférences, dont le remplissage de « formulaires d'entité ». Le formulaire permet de relier à une entité donnée un ensemble d'informations de natures diverses. En nous fondant sur les expériences menées dans le cadre de MUC, nous pouvons offrir par exemple une représentation enrichie de la notion d'*organisation*.

```
Entity{
    Lexical_unit=ONU;
    Sem{
        Type=organization;
        Focalisation=none;
    }
    EntityTemplate{
        IsLocatedIn = New_York;
        IsComposedOf = employees && diplomats;
        IsLeadedBy = Kofi_Annan;
        KindOf =diplomatic_org)
    }
}
```

Ce type de structures permet d'avoir accès aux différentes composantes de l'entité et permet d'expliquer certaines dénominations fondées sur la métonymie ou la mise en avant d'un des aspects de l'objet visé. Le modèle développé rend explicite les apparents changements de catégorie qui peuvent s'expliquer par une fonction d'accès à un des aspects de l'entité. Les outils d'extraction d'information (Poibeau, 2003) peuvent ici s'appliquer pour contribuer à produire ces ressources automatiquement. Ils demandent toutefois la mise au point de systèmes collaboratifs, dans la mesure où toute l'information ne peut pas être acquise directement à partir du corpus. Le peu de ressources disponibles pour le français, notamment au niveau sémantique, rend la tâche particulièrement difficile.

## 4.3 La question des anaphores nominales

Les mécanismes de mise en avant d'un aspect de l'entité en fonction du contexte restent à étudier. L'entité est constituée d'un ensemble de traits permettant d'avoir accès à ses différentes composantes. Il est ainsi possible d'expliquer certaines anaphores nominales qui résistent traditionnellement à l'analyse :

*L'organisation de Kofi Annan…*
*Syn(L'organisation de Kofi Annan) = ONU*

*Justification: IsLeadedBy(ONU)=Kofi Annan*

Ces mécanismes peuvent être modélisés au moyen de fonctions à la Melc'uk pour offrir une version unifiée du lexique, intégrant connaissances linguistiques et connaissances sur le monde. En ne cherchant pas à rendre explicite la référence de l'entité par rapport à un modèle du monde, notre modélisation se démarque nettement des travaux effectués dans le cadre de l'analyse du dialogue (Popescu-Belis, 1999 ; Salmon-Alt, 2001). Elle rend en revanche explicite les effets de sens au sein de l'entité, afin d'expliquer les phénomènes de synecdoque et de métonymie entrevus précédemment.

# 5  Conclusion

Nous avons essayé de montrer dans cet article qu'une même entité peut être désignée de multiples façons et que les noms désignant ces entités sont par nature polysémiques. L'analyse ne peut donc se limiter à une tentative de résolution d'anaphores mais doit mettre en évidence les possibilités de nommage s'appuyant essentiellement sur deux opérations de nature linguistique : la synecdoque et la métonymie. Nous avons enfin présenté une modélisation permettant de rendre explicite les différentes désignations en discours, en unifiant le mode de représentation des connaissances linguistiques et des connaissances sur le monde.

# 6  Références